# Spiking Neural Networks Hardware Implementations and Challenges: a Survey

MAXENCE BOUVIER, CEA-LETI, Minatec Campus
ALEXANDRE VALENTIAN, CEA-LETI, Minatec Campus
THOMAS MESQUIDA, CEA-LETI, Minatec Campus
FRANCOIS RUMMENS, CEA-LETI, Minatec Campus
MARINA REYBOZ, CEA-LETI, Minatec Campus
ELISA VIANELLO, CEA-LETI, Minatec Campus
EDITH BEIGNE, CEA-LETI, Minatec Campus

Neuromorphic computing is henceforth a major research field for both academic and industrial actors. As opposed to Von Neumann machines, brain-inspired processors aim at bringing closer the memory and the computational elements to efficiently evaluate machine-learning algorithms. Recently, Spiking Neural Networks, a generation of cognitive algorithms employing computational primitives mimicking neuron and synapse operational principles, have become an important part of deep learning. They are expected to improve the computational performance and efficiency of neural networks, but are best suited for hardware able to support their temporal dynamics. In this survey, we present the state of the art of hardware implementations of spiking neural networks and the current trends in algorithm elaboration from model selection to training mechanisms. The scope of existing solutions is extensive; we thus present the general framework and study on a case-by-case basis the relevant particularities. We describe the strategies employed to leverage the characteristics of these event-driven algorithms at the hardware level and discuss their related advantages and challenges.

## 1 INTRODUCTION

Machine learning algorithms have been a growing subject of interest for nearly 40 years. The complexity of neural networks has progressively increased, and the field is now referred to as deep learning due to the very large amount of layers composing them. We distinguish two types of neural networks: the formal ones, referred to as Artificial Neural Networks (ANNs), and the event-driven ones, also called Spiking Neural Networks (SNNs). The former have been studied since the beginning of machine learning, the latter are more recent, the first reports referring to SNNs being published around 20 years ago. Wolfgang Maass derived that their computational power is at least equal to the one of ANNs [116].

Because of the increase of the computational density of those algorithms, their evaluation on a traditional computer architecture, known as Von Neumann, now requires an extreme amount of time and power. As a result, hardware designers started working on accelerators dedicated to the evaluation of such algorithms. Neural networks are based on the biological neural computation scheme occurring in the brain, it is thus no surprise that processors dedicated to their evaluation is also taking inspiration from the brain structure. Such circuits are therefore referred to as neuromorphic [121].

Neuromorphic computing can be applied to both formal and event-driven networks. The scope of applications for machine learning is probably infinite, but their practical realization remains limited due to hardware resource requirements, especially for embedded application. Indeed, battery constrained devices usually require a connection to the cloud for external evaluation on more powerful machines. Hence, there is a need for low power hardware and algorithm co-design to enlarge the applicability of deep learning.

There is a growing belief that spiking neural networks may easily enable low power hardware evaluation. Indeed, they are supposedly efficient for both computation and communication thanks to their event-driven sparsity. However, on this day, it is not yet clear whether hardware implementation of SNNs is effectively efficient. Therefore, in this survey, we deliver an overview of the advancement in this field. We detail the working principles of SNNs; how it can be adapted to hardware and how to leverage their performance gains.

In the next section, we present the neural networks operation, both formal and spiking, and explain how neuromorphic computing differs from Von Neumann machines. The third section discusses the bio-inspiration of SNNs and how bio-plausible models can be implemented using hardware. Section 4 focuses on SNNs for machine learning applications. We review the different strategies allowing to reduce the hardware resources required to evaluate SNNs for machine learning applications, and study to what extent SNNs are suited for low power applications. We also discuss the state of the art regarding crossbar arrays implementations. The fifth section examines the ways of training a SNN and discusses the interest and feasibility of on-chip learning. The sixth section gives the state of the art of SNNs' hardware implementations, for both large scale realization and small scale targeting low power for potential embedded operation. The seventh section introduces potential paths for future improvements of SNNs, both on the software and hardware point of views.

## 2  SPIKING NEURAL NETWORKS WORKING PRINCIPLES

### 2.1  Introduction to Neural Networks

The global interest of researchers and more recently industry in artificial neural networks is based on: (1) their potential as nonlinear universal approximators [65], (2) the possibility to train them. That is, iteratively program them to realize their function with a general algorithm, namely backpropagation of the gradient, applicable to any network or target function or both, without precise prior knowledge of how the function should be described [187]. To make it simple, we describe the structure of a multi-layer perceptron, the basic Artificial Neural Network

ANNs are inspired from the working mechanism of the brain. They are a succession of computational primitives, namely neurons and synapses. Neurons, also called perceptrons, are assembled into layers and each layer is connected via a set of synapses, whose connections depend on the network topology. A representation of a biological neuron and its artificial perceptron couterpart is given in Fig. 1.

A synapse connects only one pre- and one post-synaptic neuron. A post-synaptic neuron implements the multiplication and summation of the pre-synaptic incoming activation values $x$ with their corresponding synaptic weights $w$ (also called synaptic efficacy) and adds a bias term $b$ to the result. Hence, the post-synaptic neuron evaluates:

$$y_j = \sum_i w_{i,j} \times x_i + b_j \tag{1}$$

At that point, perceptrons are linear, whereas their nonlinearity plays a crucial role for the universality of neural networks [65]. Hence, on top of that, a nonlinear activation function $f$ is applied on $y_j$. At the end, keeping the previous notation, an incoming activation value $x$ of a neuron in layer $l$ is equal to $x^l = f(y^{l-1})$.

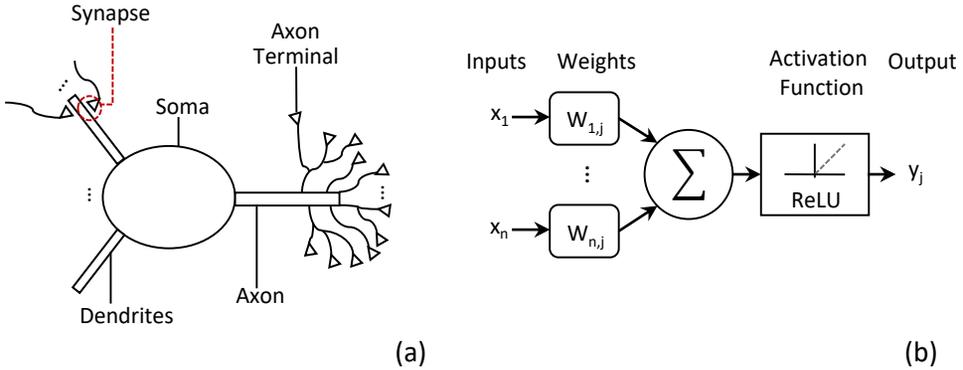

(a)                                                                                      (b)

Fig. 1. (a) Schematic representation of a biological neuron. Dendrites receive inputs from upstream neurons via the synapses. The soma membrane voltage integrates those inputs and transmits its output to an axon. Axon terminals are in charge of transmission among many downstream neurons. (b) Schematic diagram of a non-linear perceptron as described by Eq. 1 on top of which is added a non-linear activation, such as the represented Rectified Linear Unit (ReLU).

The connections between layers may have various configurations. Fully Connected (FC) networks consist in having all the units of a layer connected to every units of the adjacent layers, such as described in Fig. 2. So, if two layers of respective capacities $n_1$ and $n_2$ are fully connected (FC) the total number of synapses is $n_1 \times n_2$. Complementary information on the various topologies, their working principles, and tutorials can be found in [56,86,134,161].

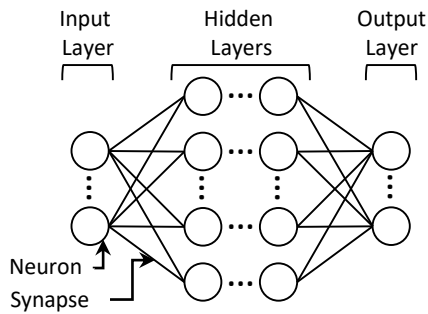

Fig. 2. Schematic representation of Fully Connected (FC) neural network.

A Spiking Neural Network (SNN), which is the purpose of this survey, follows more sophisticated design rules. One could summarize the general behavior as follows: when a synapse receives an action potential, also called a spike, from its pre-synaptic neuron, it will emit a post-synaptic potential (PSP), which stimulates the membrane of its post-synaptic neuron. The neuron membrane potential is a function of incoming synaptic stimulations and evolves with time. If the potential overcomes a threshold, the post-synaptic neuron fires, i.e. it emits an action potential taking the form of a spike. The operation of a simple integrate and fire (IF) neuron is illustrated in the inlet (c) of Fig. 3. We discuss further the different models of neurons in section 3.1.1. In [116], Wolfgang Maass argues that SNNs should allow to compete with and overcome formal (as opposed to spiking) NNs' computational power for machine learning applications. In this paper, he compares the number of units required to realize a function and demonstrates that by using spiking neurons the same amount is required for any function and may be less for specific functions. Moreover, the integration of time in the information propagation and computation of SNNs may enable them to efficiently extract temporal information from time dependent data [54].

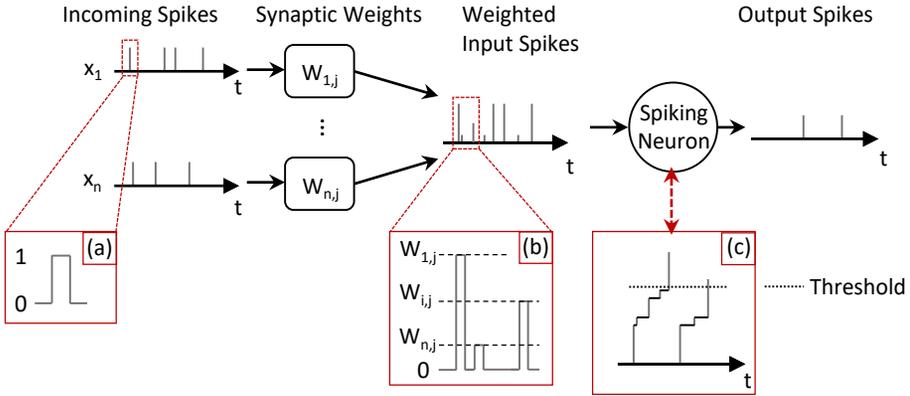

Fig. 3. Schematic representation of a spiking neuron. Incoming binary spikes, received through the synapses, are weighted and integrated. The neuron integrates them and emit in turn binary spikes to downstream units. Inlet (a) represents a binary spike. Inlet (b) illustrates the synaptic weighting of the spikes. Inlet (c) shows the integration process on the membrane potential of a simple Integrate and Fire (IF) neuron model.

To sum up, spiking neurons communicate via impulsions, which are described using binary signals, and integrate over time the incoming spikes onto their membrane. An advanced description of the neuron and synapse operations in spiking neural networks is discussed in more details in section 3.1.

## 2.2 Information Representation Using Spiking Neurons

In SNNs, a spike takes the form of a single binary bit. The information representation with binary spikes allows these systems to be faithfully implemented using analog or digital hardware. In order to code network inputs, and readout their outputs, one needs to understand how to represent complex information using spikes. However, information representation in networks of spiking neurons is still being discussed [100,142,156]. The majority of artificial SNNs use what is commonly called rate coding, whereby the information transmitted between neurons is represented as the mean firing rate of emitted spikes over an observation period. Poisson spike trains are usually employed, in which case the precise timing between consecutive spikes is random but the overall frequency of spikes is fixed. However, some neuroscientists suggest that part of the information in the brain is encoded in a temporal manner [180]. Time coding can be implemented under several forms. Some use the time to first spike (TTFS) [154], where intensity of the activation is inversely proportional to the firing delay of the neuron: the one with the highest membrane potential fires first. Others, as presented in [35,100,124], use the inter-spike interval (ISI), i.e. the precise delay between consecutive spikes, to code the intensity of the activation. Nevertheless, Fairhall et al. suggest that the encoding is realized simultaneously at several time scales [43], which suggests that real brain communication exploits a combination of the three.

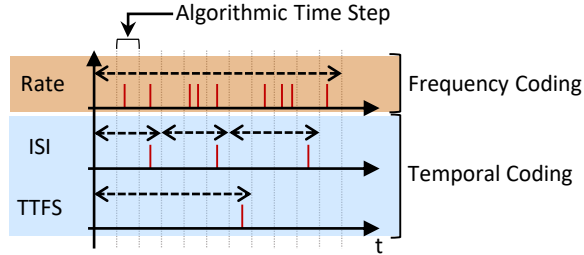

Fig. 4. Temporal diagram of the number of emitted spikes as a function of the type of coding employed.

In addition, it is also possible to use information encoding only for input conversion, from frame information to event driven information. The successive layers of neurons will then simply integrate the spike pattern they receive and fire as soon as their membrane potential overcomes their threshold.

## 2.3 General Hardware Implementation Strategy

### 2.3.1 Hardware Platform.
Simulation of SNNs is usually performed by discretizing the time, and evaluating the state of every neuron at each algorithmic time step. The shorter the time step, the more accurate the simulation is, but the larger its duration. Even if leveraging the computational sparsity of SNNs in simulation is possible, by updating only the neurons receiving spikes at each step, only a dedicated hardware is able to take full advantage of the spiking behavior. To accelerate the computation of SNNs, neuromorphic ASICs and ASIPs have recently been proposed. The major realizations are BrainScaleS from an European Consortium [160], the recent Loihi from Intel [38], Neurogrid from Stanford University [15], ROLLS from the Institute of Neuroinformatics (INI Zurich) [148], SpiNNaker from the University of Manchester [49], and TrueNorth from IBM [122].

It is largely accepted that the main limitation in performance, in terms of throughput and power consumption, when evaluating ANNs, comes from the memory bottleneck [9,66,118]. With inspiration taken from the brain computing paradigm, neuromorphic processors' ambition is thus to distribute the memory of the whole architecture within close proximity to the Processing Elements (PEs). This leads to parallelizing the storage and computation of the different layers of an ANN, while reducing the power consumption of the whole operation. It goes the same for SNNs' hardware evaluation. The computational core is separated into several small neurocores, where memory and PEs are specifically arranged. Hence, the layers of the network are distributed/mapped among the neurocores, and each neurocore both stores part of the synapse weights and performs the neuron evaluations of the SNN topology.

In input/output a core receives/transmits spikes via the Address Event Representation (AER) protocol through a communication scheme like a Network on Chip (NoC) [19,111]. This type of architecture is scalable as long as the routers and the control circuitry can manage the AER requests. Still, the definition of the number of neurons and synapses per core, and the number of core per chip will limit the implementable topology onto a chip. To overcome this limitation, some implementations even realized scalable multi-chip architecture [48].

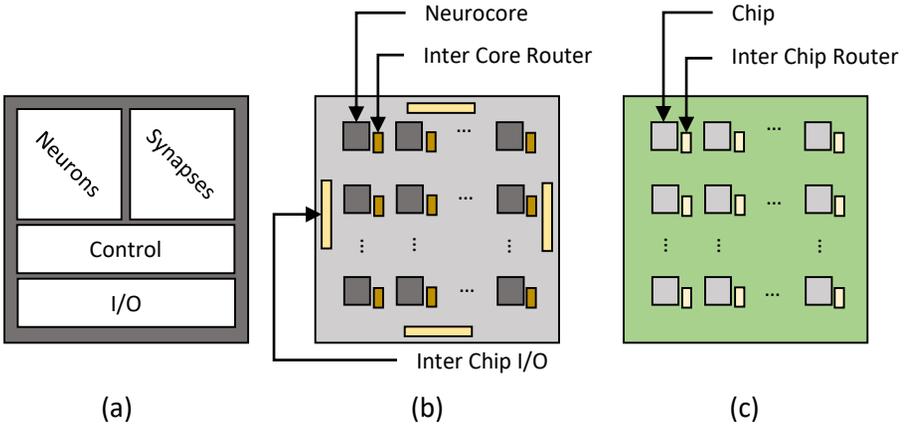

Fig. 5. General neuromorphic hardware description at different scales. (a) Neurocore representation. It integrates a local memory, algorithmic and control circuitry to evaluate the network mapping. (b) Representation of a chip, it integrates several neurocores and a NoC to communicate AER events between cores. (c) A board representation containing several neuromorphic chips.

### 2.3.2 Algorithm to Hardware Mapping.

The mapping strategy employed to evaluate a SNN onto a neuromorphic hardware may largely affect the performances of the hardware evaluation of the circuit.

An intuitive mapping strategy is to have one core evaluate one layer, as depicted in Fig. 6. (a). With this solution the computational efficiency of the neuromorphic chip is highly dependent upon the network topology [153,182]. For example, let us consider a chip composed of three identical neurocores that evaluates a FC network with a relatively small topology of 100-1000-10 units. In this case, a single spike in input of the hidden layer requires 1000 neuron updates. However, a single spike in input of the last layer requires only 10 neuron updates. Hence, if each layer receives an equivalent amount of spikes at each time step, the computation charge is not distributed among the neurocores, the one dedicated to the hidden layer being the most active. Nevertheless, if the second layer receives 100 times less spikes than the third one, the number of updates realized by both neurocores is equivalent. Then, what becomes important is to make sure that the distance travelled by the spikes between neurocores is minimal. Therefore, depending on the layer organization among the available neurocores, their relative activities vary, as well as the number of packets sent through the NoC. In the example, we dealt with a relatively small network, but current deep networks can contain up to a million units, which ultimately increases the number of neurocores required to compute the full network, as well as the neurocore specifications in terms of both memory and computational capabilities. The first approach may thus not be an ideal solution.

A second mapping strategy would be to implement parallelization along the spatial dimension (as opposed to the depth) of the network. In this configuration, each neurocore will compute a sub-part of a NN layer, as described in Fig. 6 (b). That solution has the advantage of enabling per-layer time mutliplexing of the whole NN computation [2], thereby reducing the required number of neurocores to compute deep networks. Nevertheless, in the case of a Convolutional Neural Network [101], every neuron of a single layer shares the same weight values. This thus leads to memory redundancy between the neurocores, and ultimately to a loss of efficiency.

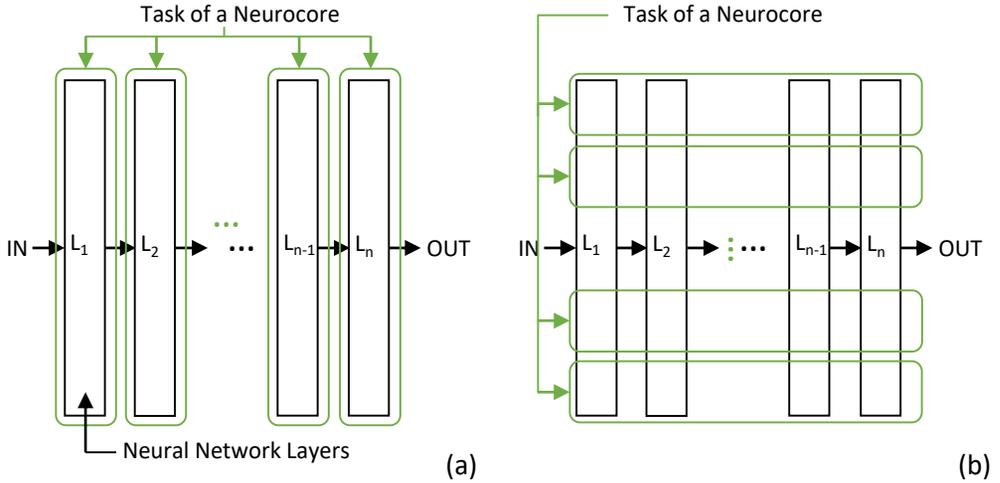

Fig. 6. Illustration of two different algorithm to neuromorphic hardware mapping strategies. (a) Parallelization along the depth of a neural network, in which case each neurocore computes a single layer. (b) Parallelization along the spatial dimensions of a neural network, in this case each neurocore evaluates a subpart of every layers.

This short case study reveals that correctly mapping the algorithm to the hardware is essential to fully exploit neuromorphic hardware capabilities. In real situations, researchers exploit a mix of both strategies to uniformly distribute the neurocores activity. Some groups are thus working on algorithms to perform optimized mapping of any trained network onto neuromorphic hardware [37,50]. Others exploit graph theory to maximise parallelization at the computation diagram level [2,173].

We will discuss further accelerators implementation details and challenges in section 4.

## 3 BRAIN INSPIRED SPIKING NEURAL NETWORKS

In the early days of neuromorphic computing, the goal was to develop novel kind of circuitry, which closely mimics the brain. The human brain consumes around 25W for 86 billion neurons [63]. Hence neuromorphic computing paradigm was born.

We differentiate two purposes of brain-inspired computing. One is to leverage neuroscience discoveries to realize low power machine learning or to improve performances of existing systems such as fault tolerant hardware. Another is to implement on-chip bio-plausible dynamical elements to either accelerate neuroscience research or interact smoothly with real biological systems. The former is studied in section 4, while in this part, we discuss some mechanisms observed in the brain which inspired spiking neural networks, event-driven circuits and sensors, excluding brain-inspired learning rules which we discuss in section 5.2. We detail brain-inspired neural networks and electronic circuits, and discuss applications that ensue from it.

### 3.1 Computation in the Brain

The brain characteristics are different from traditional computers [206]. In short, it is highly parallel with interwoven computation and memory elements, deployed in volume, and operates asynchronously.

*3.1.1 Neuron Models.* Neuromorphic systems, more precisely networks of spiking neurons, are designed following a bottom-up approach. Building blocks, namely neurons and synapses, are designed and assembled into networks. The neurons are the computing elements of neuromorphic circuits. Several models describe the neuron functions, with a varying degree of complexity [52,53,75]. A large amount of them have already been transposed onto hardware [11,20,31,73,198]. The Leaky Integrate and Fire (LIF) model has encountered a special interest among hardware designers [1,15,38,122], we thus discuss its working principle below.

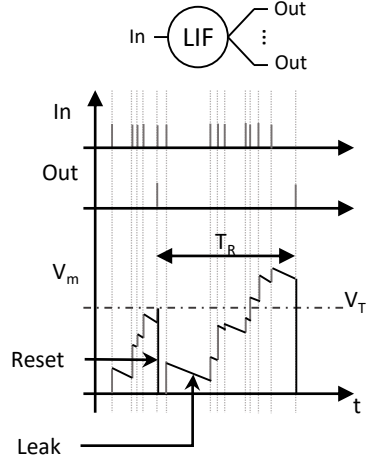

Fig. 7. Membrane potential temporal evolution of a typical leaky integrate and fire neuron. Several spikes are received asynchronously, the reset sets back the potential to zero and the refractory period forbids the neuron to spike until enough time has passed between two spikes.

The temporal activity of a usual LIF neuron is depicted in Fig. 7, and its behavior is described as follows. When a spike inputs the neuron, i.e. $x_i(t-1) = 1$, the synaptic weight $w_i$ associated to this spike will be integrated on the membrane. When the membrane potential $V_m$ overcomes a threshold $V_t$, the neuron fires, i.e. $y_i = 1$ and resets its membrane potential. Nevertheless, if the refractory time $T_R$ is not reached, i.e. the amount of time since the last output spike is smaller than $T_R$, the neuron does not fire, even if its membrane potential is above the threshold. In addition, due to leakage, the membrane potential decreases continuously at the leak rate between two input spikes.

The essential parameters of a LIF neuron are the membrane threshold voltage, the reset potential, the refractory period and the leak rate. At each time step $t$, the membrane potential of neuron $j$ of layer $l$ can be described by

$$Vm_j^l(t) = Vm_j^l(t-1) + \sum_i w_{i,j} \times x_i^{l-1}(t-1) - \lambda \qquad (2)$$

Where the $\lambda$ parameter corresponds to the leak, and $w_{i,j}$ is the synaptic potentiation. Note that models including complex internal mechanisms may be used to define these parameters, instead of constant scalar values.

After integration of all incoming spikes for a given time step, the potential is compared to the threshold and the output is defined by

$$\begin{cases} x_j^l(t) = 1 \; if \; Vm_j^l(t) > V_t \; and \; t - t_{spike} > T_r \\ \qquad x_j^l(t) = 0 \; otherwise \end{cases} \qquad (3)$$

Where $t_{spike}$ is the last time step at which the neuron $j$ of layer $l$ fired. Reset can be performed following different schemes: reset the potential to a constant value $V_r$ or subtract the reset value from the current membrane potential. Rueckauer et al. [155] discuss the effect of both on final network accuracy. These units are strongly nonlinear elements in themselves because their output is a succession in time of binary spikes.

The LIF neuron model is already a simple one with respect to biological dynamics [116], but it is not the simplest. Designers that do not target bio-plausible tasks can use abstracted models to reduce hardware resources, such as the Integrate and Fire (IF) neuron, a subset of LIF whose refractory and leakage mechanisms are removed, such as described in Fig. 3. The necessary electronic building blocks to implement it are an adder, for integration, a comparator, for threshold detection, and a memory, for membrane potential storage [15].

On the other hand, the bio mimicking is sometimes pushed further, with the hardware implementation of ionic channels and other bio-realistic components [15,70,148,191]. These biomimicking circuits allow to emulate close-to-real dynamics and are expected to deliver new insights on neuronal function from the neuroscientific point of view. Biomimicry hardware can be realized using both digital and analog electronics design. We have seen some groups [31] implementing advanced reconfigurable units based on the Izhikevich work [75] or bio-realistic ion channels [191] interaction in fully digital designs. SpiNNaker enables evaluation of complex neuron and synapse models but leads to heavy computational tasks [49]. Recently, Partzsch et al. proposed an exponential approximation accelerator to improve the performance of SpiNNaker neuromorphic processor [139]. Nevertheless, most of the biomimicking hardware has been realized using analog electronics. It is mostly due to the fact that complex mechanisms can be implemented with transistors biased in the subthreshold regime, where exponential behaviors are commonly observed [121]. Two major analog emulators are Neurogrid [15] and ROLLS [148], enabling real time (non discretized) and close to real neural simulation.

*3.1.2  Synapse Models.* Emitted spikes transit through synapses before reaching downstream neurons. ANNs consider the synapses as scalar values, the weight, which is different from the brain. A biological synapse receives spikes and, in turn, stimulates the membrane potential of post-synaptic neurons via a current, whose value evolves during time [44]. The synapse's weight, as described in formal networks, actually represents the Long Term Potentiation (LTP) phenomenon. The intensity of the excitatory current can reach various levels, which is determined by past long term activity of the synapse. Additionally, another mechanism, called Short Term Potentiation (STP), modulates the excitatory level according to the recent activity of the synapse [186]. It reduces the synapse's output intensity of each incoming spikes arriving one after the other in a short period of time. Again, Neurogrid [15] and ROLLS [148] realize such synaptic behaviors.

*3.1.3  Other Computational Elements.* In addition to complex computational primitives, their organization and relations in the brain lead to higher-level mechanisms for performing operations. Software and hardware designers have taken inspiration from it to improve performance of deep learning algorithms in a top-down approach.

Convolutional Neural Networks (CNNs) are probably the best illustration of it [101]. They are inspired from the hierarchical layering for visual recognition observed in the brain [68,165], and have largely contributed to enabling image classification algorithms to reach human performance [157].

Winner-Take-All (WTA) [117] is another brain-inspired mechanism implemented in inhibitory neural networks. It consists in having the units of a single layer compete with each other, where the one receiving the strongest input spikes and inhibits all the others. A variant is the kWTA, where each layer can have *k* firing neurons before full inhibition of the layer. WTA is a major factor of decision-making [4,159].

Neuron firing synchrony has been observed in the visual cortex and has been associated with dynamic interactions within the network [168]. Hence not only is information encoded by individual neurons with rate or time coding, it is also communicated on a population level by the simultaneous activity of different units. We argue that artificial SNNs could capitalize on this observation to enhance their computational capability, especially for datasets where inputs timing is relevant to the targeted application.

Finally, a specific type of spiking networks, called liquid state machines (LSMs) or reservoir networks, fully takes its inspiration from the brain. Their topology is unlike traditional FC or convolutional networks, in the sense that there are not organized in layers. They are a combination of neurons randomly interconnected where each unit can receive spikes from both the input and the other neurons. Because of their recurrent nature, they are usually used for time dependent applications such as speech recognition. Given their complex computational graph representation, they may not map efficiently onto general neuromorphic hardware. Hence, some research groups are concentrating their efforts on designing neuromorphic architecture specialized for LSMs on FPGA [80,162,185] as well as other hardware support such as photonics [137] or 3D ICs [98].

To sum up, spiking neural networks implement neuron and synapse models with various degrees of complexity, as well as possible higher-level brain inspired functionalities. Complex networks require specific hardware to compute efficiently their evolution through time. Analog electronics can faithfully implement advanced neuromorphic operations, especially in the subthreshold regime, but digital realizations of bio-plausible mechanisms have also been proposed. Nevertheless, the use of bio-plausible models is not required for every application relying on SNNs. Machine learning may rely on simplified spiking units, such as LIF or IF models, to enable accurate and efficient hardware implementations.

## 3.2 Brain-Inspired Technologies

*3.2.1 Autonomous and Robust Systems.* The scope of applications resulting from bio-inspiration regarding fault tolerance and adaptation mechanisms is large.

For what concerns adaptation, a famous resulting field is machine learning. Closer to the hardware, software training can also be used to adapt the program onto defective hardware, for example high variability between novel nanodevices can be autonomously compensated for [108].

Several strategies for developing resilient electronics have emerged, based on the impressive resilience against noise, fault or damage of biological systems. For instance, the principles of evolution and natural selection gave rise to evolutionary hardware for implementation of reconfigurable fault tolerant systems [10,58,151]. Closer to the brain, we have seen researchers exploit neuroplasticity mechanisms to realize efficient fault tolerant and reconfigurable systems [83,109,110,166]. Neuroplasticity refers to the ability of the brain to adapt and reconfigure during lifetime operation. It involves connectional structure modification, synapse adaptation, and others [143]. Its implementation onto hardware is thus relatively complex, but may allow to remove the need for redundant hardware in critical electronic systems. Some even take inspiration from higher-level brain mechanisms, such as auto-encoder networks, to realize noise resilient circuits [90].

Hence, bio-inspired computing is not limited to machine learning, and already benefits to the implementation of fault tolerant systems.

*3.2.2 Event Driven Sensors.* On top of circuits whose organization and computation schemes are bio-inspired, sensors mimicking the biological senses have also been implemented. Among them are dynamic vision sensors [107,145] and audio recording systems such as the cochlea circuit [32]. They output the data in the form of spikes, using the AER protocol, and their sensing principles are based on bio plausible mechanisms. Their main advantages over frame based sensors are their fast response time and important dynamic range [107]. Detailed information on their implementations and working principles can be found in [105,111].

## 4   NEUROMORPHIC MACHINE LEARNING ACCELERATORS

Neuromorphic systems for machine learning applications have recently become a major subject in computer design, for both ANN and SNN acceleration. ANNs evaluation being computational intensive, hardware for fast and low power evaluation of ANNs, such as the Tensor Processing Unit [85] are currently being developed. In parallel, SNN accelerators are becoming essential for the evaluation of spiking neural networks. Indeed, their simulation requires sequential computation of several time steps, and simulation time thus scales with the deepness of the network as well as the time resolution of the simulation. As discussed in section 3, hardware implementation of SNNs enables the emulation of more or less bio-plausible neurons and synapse models, which may help acquiring more knowledge of the operation of the brain. Machine learning application is also done with SNNs but their simulation takes a large amount of time on traditional computers because of the time discretization. To such use, SNN hardware implementation may leverage properties of SNNs to reduce power consumption and increase evaluation speed, especially when using hardware friendly models of neurons and synapses, such as the simple IF neuron and constant weights.

We discuss here the neurocore organization in details, the communication scheme between cores with binary spikes, the sparsity of SNNs, how the fault tolerance of neural networks can be taken advantage of, and finally the implementation of SNNs with crossbar arrays.

### 4.1   Core Organization for Low Power Spiking Neural Network Evaluation

As discussed is section 2.3, a neuromorphic processor is a chip composed of one or several neurocores. A single core integrates Processing Elements, to evaluate the neuron membrane potential, memory to store synaptic values and neuron states, input and output interfaces, to receive and emit spikes, and control circuitry. The number of neurons computed per neurocore relies on the size in memory of the parameters required per unit. Designs targeting very large scale applications usually implement a large number of neurocores to maximize parallelization at chip level [15,38,49,122,148,160]. A large amount of publications report various neurocore organizations with designs optimized in accordance with the characteristics of the network topology to be mapped to the hardware [28,91,130,164,193], or with a targeted application [34,38,103,148,153,192,194,201].

Different network topologies require different hardware designs to fully take advantage of neuromorphic computing. For example, a shallow fully connected network does not have the same algorithmic footprint as a deep convolutional neural network, nor the same memory requirements [181]. However, given the large spectra of existing network topologies and their constant evolution, we classify neurocore designs following a hardware-wise, non network-specific distinction: analog or digital design. To implement the processing elements and memory using analog or digital electronics is a designer choice based on the targeted applications and performances. The goal of neuromorphic implementations may be to emulate precise bio-inspired neuronal computation, to compete for the lowest power or highest accuracy on a specific machine learning benchmark, or to enable very large-scale neural simulation at low cost, both in time and power.

In section 3, we discussed analog implementation and concluded that it is well suited for the realization of complex bio-plausible tasks [148,191]. Nevertheless, machine-learning using such circuit designs is still possible, their main advantage being the ideal co-localization of memory and computation [15,72]. Qiao N. and Indiveri G. [147] recently reported hardware for both real-world interfacing or SNN acceleration, using mixed signal digital-analog circuits whose speed can be adapted to application requirements.

Reported digital designs usually employ a single PE which updates every neurons mapped to the core [28,34,38,122,130,153,167,193,194]. For each algorithmic time step, the neurocore runs through the following procedure: first, every upstream spikes received during the previous time step, which are stored in an event queue such as spike schedulers [153,193] or FIFO registers [28,115], are processed sequentially. Each spike is an address which allows to identify the downstream neurons according to the connection map of the part of the SNN processed by the neurocore. Thus, every synaptic weights of downstream neurons as well as neuron state parameters are loaded from the core memory. The update of each neuron is then done sequentially, in a time-multiplexed manner. A neuron update consists in adding the value of the weight associated to an input spike to the neuron's membrane potential, after what the membrane potential is compared to the threshold of the neuron, as depicted in Fig. 3. As soon as a neuron's membrane potential crosses its threshold, an output spike is emitted onto an asynchronous network, its membrane is reset, and the neurocore operation proceeds until every spikes from the previous time steps are processed. One interest of spiking neural networks is thus that an operation is reduced to an addition and a comparison [193]. Hence, digital implementation only requires adders, which significantly reduces computation cost when compared to MAC operators of ANNs. However, because time is a primitive in spiking neural networks computation [116], both timing information (for on-chip learning, leakage and refractory period mechanisms) and neuron state variables (membrane potential, threshold) have to be stored, which requires supplementary memory [38,74,201]. Depending on the network topology, weight access quickly becomes a bottleneck given the possible large number of units and connections; there are around 10,000 synapses per neuron in deep networks. To reduce the memory impact, several strategies can be followed: its distribution inside neurocores, the use of hierarchical memory to pre-load data, and the employment of new kind of Non Volatile Memory (NVM) devices [25].

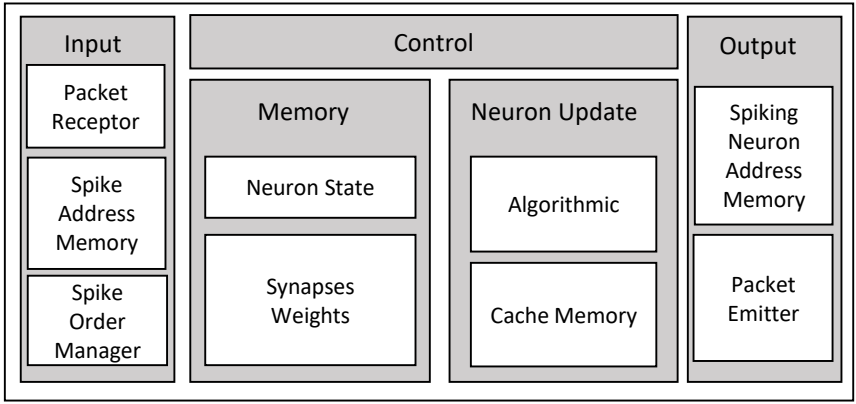

Fig. 8. Schematic representation of the neurocore components.

A problem of time multiplexing the neuron updates is that a single algorithmic time step takes several clock cycles to complete. The more the neuron model is complex, with leakage, refractory period, etc., the more the number of clock cycles will be important. Hence, some groups propose to smartly reduce computation cost by updating only what is necessary. For example, Roy et al. [153] implemented LIF neurons, but realized "leak-upon-load", so that leakage operation was not realized every time step for all neurons. On the other hand, it requires to store the last time the neuron was loaded. Shihui et al. [193] compressed their network, resulting in a small number of neurons per layers. They thus parallelize unit updates and remove time multiplexing requirements enabling them to perform voltage and frequency scaling to reduce power consumption while keeping their targeted throughput. Note that if speed is not an issue, voltage and frequency scaling has also been used without network compression [64,96]. In [115], authors physically implement a few neurons and still perform time multiplexing; as such, they create a tradeoff between hardware requirements and throughput reduction.

Many other design choices can help to improve performances. The synchronous operation but asynchronous communication between neurocores requires the presence of event queues to receive the incoming spikes asynchronously during an algorithmic time step. As such, spike schedulers [153,193] are powerful tools. They allow organizing the sequence of upstream spikes to be delivered to downstream neurocores in order to reduce potential redundant memory accesses. Roy et al. [153] also interestingly share them between several neurocores. This may maximize the efficiency of topology mapping, for instance by allowing a single spike scheduler per SNN layer, even for networks with large layers.

At a higher scale, neuromorphic design, taking inspiration from the parallel, highly distributed brain architecture, are commonly composed of several neurocores, communicating together via a NoC [48,74]. Very large-scale designs even implemented multi-chip boards for evaluation of deep networks [48]. This parallelization at every scale is possible thanks to the scalability of the AER protocol. It allows distributing the computing task in several smaller computations and by doing so accelerates the overall network evaluation. Moreover, one could program the chip so that each SNN layer is computed within one neurocore or cluster of neurocores spatially close together on the network, to pipeline the overall network evaluation and fully take advantage of the bio-inspired hardware design.

To sum up, neuromorphic design allows to implement parallel SNN evaluation. Given the amount of properties that contribute to power consumption, throughput, and chip size, we argue that accelerators must respond to their target application. A processor designated to training or evaluation or both of very large scale SNNs should maximize flexibility and speed, leaving besides power consumption and chip size. However, an accelerator designed to target embedded applications should focus on power consumption and area requirements, and be designed in close consideration of the SNN algorithm to be evaluated. The neurocore architecture is an essential component, but other tools, such as the AER protocol and fault tolerance exploitation, improve neuromorphic chip performances.

## 4.2 Reduced Communication Cost

Multicore parallelism of SNN accelerators relies on a specific NoC communication protocol to transmit events between a large number of neurocores at minimal power and delay. The Address Event Representation is always proposed in neuromorphic systems. It enables to naturally encode the time and organize the connectivity at low communication cost [19,111]. It consists in sending a unique packet containing the address of the spiking neuron on a digital bus with asynchronous logic. As soon as a neuron fires, its address is sent onto the NoC, and the firing time is encoded in real-time on the asynchronous bus.

When compared to the frame approach of ANNs, the benefits of the AER protocol for large scale SNNs computing are numerous [111]. First, Boahen [19] has shown that it allows to reduce the network bus size while conserving a large connectivity capacity. AER NoC area requirements are thus minimum, and enable large-scale design. In addition, the transmitted packets being reduced to a single address, it also guarantees small delay and power (due to switching activity) overheads. Moreover, the neuron activity sparsity of SNN trims the NoC activity, reducing the number of packets sent on the network.

This communication protocol is easily scalable. As long as the routers can manage the requests, any number of units can be connected together. It is thus suitable for multi-chip implementation as well. However, the power consumption of such a system grows with the number of units to connect, the mean firing frequency of each unit, and the overall chip size [37,138]. Some works target to reduce the power and delay overheads of AER by exploiting the locality and clustering of NNs algorithms [19,113]. Another possibility is to employ hierarchized router levels, which helps reducing power impact of large-scale systems [72,113,138]. Rate or time coding have significantly different impacts on such overheads. Using rate coding, a larger amount of spikes, with respect to time coding, is sent through the network and the overall system power consumption drastically increases, as shown in [130]. On the other hand, rate coding usually employs Poisson spike trains and an individual spike timing error is thus of small consequence for the precision of the network. It is different for time coding, where timing matters. Timing error or information loss may occur when the algorithmic time step duration does not allow the neurocore to complete their operation. In this situation, spikes may be missing or even dropped from the network. Nonetheless, it can be used as a tradeoff between accuracy and throughput or power [34].

Finally, the AER protocol is compatible with various NoC routing architectures and broadcast schemes, such as 2D mesh [38,49,122], multicast tree [15], or systolic ring [91,96,153]. It is thus possible for the hardware designer to adapt the routing scheme employed to best fit the requirements of the accelerator.

Therefore, the AER protocol allows low power and footprint NoC designs. The packets being reduced to a single address, the communication is very efficient. AER can be adapted on several types of NoC, with various broadcast schemes. Also, thanks to the intrinsic important sparsity of SNNs, only the necessary information is transmitted.

## 4.3 Leverage High Sparsity of Dataflow

Many argue that the main advantage of SNNs is that for a given input data, due to the event driven nature and the thresholding of every units, the evaluation will usually not require every neuron of every layer to fire [130]. As such, computation is sparse and the evaluation of the network on the input data requires less operations compared to the ANNs' frame-driven approach, where evaluation of every neurons of the network is required for every input data. Here, we describe the sparse operation of SNNs and discuss whether it advantages them over ANNs.

*4.3.1 On the sparsity of SNNs.* ANNs accelerators employ guarding strategies to exploit the sparsity of the network activations [127,189], which requires supplementary hardware to check the nullity of the data. On the other hand, when evaluating a SNN on event driven hardware, the sparsity inherently reduces the switching activity. Indeed, due to the thresholding operation, a neuron with a small activation does not fire.

Moreover, if an event-driven sensor generates the input, the number of computation is adapted to the reality requirement [22,107,145]. Event-driven sensors are fully dynamic sensors that emit spikes to encode the detection of a sensed phenomenon. They are opposed to frame-driven sensors, which deliver a fixed amount of data at a specific sampling frequency. Hence, for a dynamic vision sensor (DVS), only the portion of the frame where some movement occurs communicates data, which is less power consuming than a traditional frame-driven sensor [107].

However, SNNs do not require an event-driven sensor input to perform. If we want to benchmark a SNN against a conventional formal dataset (e.g. MNIST [101]), the dataset first requires a conversion to spiking information. For visual applications, presented solutions are: (1) conversion of intensity to Poisson spike train following an arbitrary conversion scale [29,39], (2) applying a continuous analog value (voltage or current) on the input neurons' membrane potential, which will be continuously integrated [155], (3) emit a single spike per pixel whose latency is inversely proportional to the intensity [88,154], (4) observe the dataset with a dynamic sensor to convert it into spikes [135]. Depending on the chosen conversion model, energy consumption and accuracy statistics will vary.

At the output, we also need to understand the information delivered by the spiking units, it is thus necessary to set an output evaluation scheme [154]. Depending on the information-coding scheme, several methods to decide which output neuron represents the output are available. For example, to implement an image classifier using time to first spike coding, the class can be represented by the first unit to spike, and, using rate coding, by the neuron with the highest firing rate, or the one with the highest membrane potential. Some works even add a classification layer at the output to make a decision [202].

*4.3.2   SNN and ANN comparison.* An important ongoing question with SNNs processors is their relative performance to ANNs accelerators. A major issue when comparing ANN and SNN comes from the input data conversion [198]. Indeed, a SNN requires a number of algorithmic time steps to evaluate the output, during which the input should be sent continuously. Thus, the feedforward operation of the spiking network has to be realized several times whereas the formal one is evaluated only once. Under these conditions, [59] have shown that the more complex is the dataset, the more the energy per classification of the SNN increases with respect to its formal counterpart. However, one may consider that, over the same period of time, an ANN would have computed the output a certain number of times depending on the sampling frequency of the input sensor. In this case, the ANN would be disadvantaged with respect to the SNN because the total number of MAC operations realized by the ANN is proportional to the number of times the ANN evaluates its output. We know that iso-accuracy between a SNN and an ANN on MNIST is possible for a similar network topology [155], but no information is given concerning the relative number of operations. Also, to the best of our knowledge, employing SNNs on more complex formal datasets such as CIFAR-10 or ImageNet delivers worse accuracies than a formal evaluation [155]. Hence, so far, SNN classification on formal visual datasets delivers lower accuracy results than their formal counterparts.

On the other hand, if one wants to evaluate a formal network on an event-driven dataset, it is necessary to convert continuous spiking information to a frame, or a succession of frames. Lee et al. [104] performed the conversion of the event driven N-MNIST dataset [135] by combining the spikes during one tenth of the full event stream. Hence, for each input stream of 300ms, they had 10 frames representing the events over 30ms. They trained an SNN and an ANN with the same topology on N-MNIST and converted N-MNIST respectively. In spiking, they achieved 98.66% accuracy on digit recognition whereas their formal network reached 97.8%. Moreover, their SNN allows this 0.8% difference in accuracy for 4× fewer operations. In this case, it appears than SNN performs better than ANN.

So, spiking hardware is intrinsically sparse in its activity which makes it an ideal candidate for low power machine learning applications. However, to compare the performances of SNNs to the ones of usual ANNs is not trivial because it requires the conversion of datasets, which biases the results. In short, SNNs achieve better performances, both in terms of accuracy and computation cost on event driven datasets, whereas ANNs achieve better ones on formal datasets.

## 4.4   Exploit Fault Tolerance of Neural Networks

Approximate computing has been used for a broad range of applications to reduce the energy cost of computation in hardware [60]. Two main approximation strategies are used with neural network applications, namely network compression and classical approximate computing.

Due to the large amount of parameters representing the neural networks, researchers targeting embedded application started to reduce the weight and activation maps precision to decrease the memory footprint of ANNs, method called network compression or quantization. Thanks to the fault tolerance of neural networks, as well as their ability to compensate for approximation while training, reduced bit precision entails only a small loss in accuracy [36,61,67,149]. Once transposed to hardware, weight quantization has shown 1.5 to 2.0× gains in energy with an accuracy loss of less than 1% [127,189]. An aggressive quantization to binary neural networks (BNNs) allows to use XNORs instead of the traditional costly MACs [149]. An interesting implementation leveraging both the parallel design of the crossbar array and the XNOR-net implementation is realized in [174]. They report 98.43% accuracy on MNIST with a simple BNN. Quantization is thus a powerful tool to improve energy efficiency and memory requirements of ANN accelerators, with limited accuracy loss.

These methods can be used for SNNs as well [79,150]. Indeed, [150] reports from 2.2× up to 3.1× gain in energy, and even more in area, for an accuracy loss of around 3%. What is interesting with weight quantization is that the designer can realize a trade-off between the accuracy of the SNN application against energy and area requirements of the neural networks. Approximate computing can also be achieved at the neuron level, where insignificant units are deactivated to reduce computation cost of evaluating SNNs [163].

Moreover, such computation skipping can be implemented at the synapse level in a random manner. Indeed, training ANNs with stochastic synapses enables a better generalization, which results in a better accuracy on the test sets [172,183]. Again, this method is compatible with SNNs, and has been tested both during training [133,171], operation [23], and even to define the connectivity between layers [14,34]. FPGA [167] and hardware [77] implementations of spiking neuromorphic systems with synaptic stochasticity shows that it allows to increase the final accuracy of the networks while reducing memory requirements. On top of that, nanoelectronic devices with intrinsic cycle-to-cycle variability, such as memristors [97] or $VO_2$ [77], allow to reduce area and power consumption overhead of random number generation. Chen et al. [34] also leverage probabilistic rewiring to increase their throughput. Less synapses implies less pulses to integrate and the algorithmic time step can thus be increased. Doing so, they report an acceleration of 8×, with an energy gain of 7.3× for an accuracy loss of only 0.25% on MNIST digit recognition, from 98.15% to 97.9%. Stochastic and quantized synapses can thus drastically reduce memory requirement and power consumption of SNNs accelerators, even more with pruning insignificant weights.

The other approach consists in designing processing elements that approximate their computation by employing modified algorithmic logic units [60]. Kim et al. [94] have shown that using carry skip adders enable speed and energy gains of 2.4× and 43% respectively when evaluating spiking neural networks onto neuromorphic hardware for character recognition, with an accuracy loss of only 0.97%.

Thus, approximate computing methods, both at the software and hardware levels can enable important gains in power consumption and speed. However, as the complexity of the dataset increases, along with the depth of the network topology, for example using the ResNet [62] on ImageNet [157], the accuracy loss becomes more important and may not be negligible any more [149], especially for critical application such as autonomous vehicles. It is thus unsure whether network compression technics and approximate computing are scalable and applicable to any task.

## 4.5 Crossbar Array with Memristive Devices

Neuron connections inside the brain are organized in three dimensions, which allows very dense and highly parallel networks at the smallest scale. Integration of neural networks on Silicon cannot reach similar density, being mostly two dimensional. Nevertheless, many groups are working on the most parallel implementation: the crossbar array [7,24,25,71,205]. Such design aims at combining the memory and neuron update parts of a neurocore (see Fig. 8.) inside a single unit, thus enabling speed and energy gains [5,7] and truly implementing non Von Neumann computing. It consists in having two metal lines crossing one another in an orthogonal fashion (see Fig. 9.); with a nanoelectronic device mimicking the synaptic behavior set at each cross point intersection. One direction represents the output of the pre-synaptic neurons, the other direction represents the connected postsynaptic neurons. As such, the operation of the analog crossbar array consists in applying voltages over the input lines, and reading the current of the corresponding output line. With each device's conductance (the inverse of its resistance) symbolizing the synaptic weight of a connection, the resulting currents add up together following the Kirchhoff laws, and the dot product is realized [25].

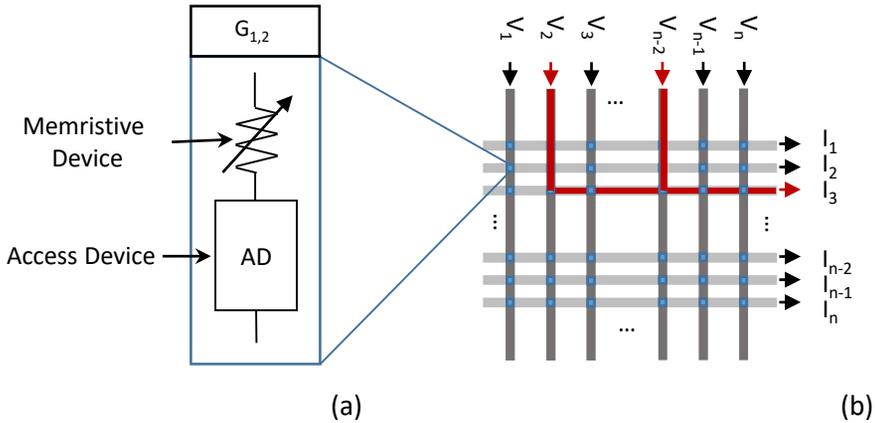

Fig. 9. Representation of a crossbar array implementation. (a) Description of a memristive device implementation. (b) Top view of a typical crossbar design, with input along the vertical lines and output along the horizontal ones.

To enable high-density crossbar implementation, without accuracy loss, the devices have to (1) be small, (2) require low power for read and write operations and (3) be stable [25]. PCM [92,175] and metal oxide resistive devices [51,119,146] are good candidates to (1) and (2) because their power consumption decreases along with their size. However, at small scale, they do not meet the third requirement yet, which drastically limits their use as neural network algorithm accelerators. Their imperfections, namely weight update non-linearity, different $G_{min}$ and $G_{max}$ values, different conductance variations for identical programming input, resistance drift, and others, limit the accuracy of a neural network mapped onto it [26,76,93]. Hence several groups focus on making better memristive elements [175] to fulfill the requirements, or on adapting the circuits architectures to compensate for the imperfections of the devices [17,26,176]. For more details regarding the working principle of the memristive elements and other types of novel devices for synaptic application, we report the reader to [25] and [11]. Some groups foresee emulating synaptic behaviors with magnetic micro and nano devices with as examples spintronic memristors [136,184] or magnetic tunnel junction synapses [171]. Up to now however, most studies are presenting results obtained from simulations with theoretical models of the devices.

The crossbar array implementation is still at the state of research because of memristive element issues, but recent papers already demonstrate impressive results. Ankit et al. [7] realized a SNN simulation at the layout level of a full neuromorphic architecture. They announce large gains, both in terms of energy and speed, when employing the crossbar array with respect to the same neuromorphic architecture designed with classical neurocores. They reach gains of several hundreds, when simulating FC networks, and several tenths when evaluating CNNs, which confirms that the crossbar gain is highly dependent on the network topology. More recently, Ambrogio et al. [5] claim reaching equivalent accuracy on ANN evaluation, with respect to an ideal simulation, using a PCM crossbar array externally controlled by a computer via a prober. They announce a potential power efficiency increase of 1 to 2 orders of magnitude with respect to standard CPUs or GPUs.

So far, what we said applies for both formal and spiking models. Nevertheless, an advantage of SNNs over ANNs is the fact that the activations are binary. Indeed, the voltage applied at each input of the crossbar is thus the same, which simplifies the surrounding circuitry. The last step, neuron update, can be realized in analog electronics, or with digital circuitry. An analog implementation can reach very high throughput and very dense implementation, ideally with a capacitor at the output of each line [92], especially if the memristive elements are directly placed between two access lines. However, because of lack of control with cross-point designs [106], crossbars are usually implemented with access devices, which drastically reduces the design density. A digital implementation of read-out circuitry will require at least an A/D converter and memory to store the neuron states [95,123]. Neuron updates can once again be time multiplexed to reduce hardware requirement, in which case the crossbar array may still be advantageous in that the computation happens in memory, but the designer must then be careful regarding the costs of transfers to and from neuron state memory and the A/D conversion. On the other hand, ANN operation with full precision activations on such designs require either simplification of the network topology [177], or advanced circuitry to deliver accurate voltage values to each input lines [3], and to apply non linearity after the MAC operation.

To conclude, crossbar array with NVM devices are promising in terms of performances and scalability, especially in the case of a fully analog implementation, where neuron update is realized in a parallel fashion. It already shows good results, but still require improvements to guarantee reliable, fully on-chip, and lasting operation life.

## 5 The Difficulty of Training a Spiking Neural Network

For a long time, retro-propagation of the gradient in SNN was not realizable due to the non-continuity of the equations of spiking neurons. Also, training a SNN requires further parameters optimization with respect to an ANN due to the fact that thresholds, leak rates, etc. are sensitive values. In order to train SNNs, several solutions were thus proposed: converting a trained ANN into an SNN, performing bio-inspired unsupervised learning, or developing some other supervised algorithms.

### 5.1 Study of Conversion ANN to SNN

At first, ANN to SNN conversion was not trivial because spiking neurons behave quite differently than formal perceptrons with nonlinear activation. Several works proposed conversion frameworks as explained in [29,39,40,141,155]. For long, the conversion from ANN to SNN required to modify the topology of the formal network [29,39,141]. However, it is now possible to design SNNs that benefit from technics such as batch normalization, inception layers, and others [155]. Thereby, Rueckauer et al. [155] achieve the best results in terms of accuracy on usual visual benchmarks using spiking networks, but are still behind the state of the art with respect to formal networks.

These transduction technics usually implement rate coding to transmit activation map values. Such coding reduces the power efficiency once the network is implemented onto hardware. Moreover, the efficiency loss scales with the dynamic range, which is rather large due to floating point representation used in conventional ANNs. To deal with this issue, Rueckauer et al. [154] have recently implemented a conversion using time coding neurons. However, this lower power implementation is at the expense of final accuracy as well as applicability of the conversion method to various ANN topologies.

As discussed in section 4.3, the difference in accuracy between ANNs and SNNs may come from the incompatibility between the datasets and the spiking operation. Nevertheless, one cannot rely only on formal to event-driven conversion to enable SNN training, especially for event-driven datasets, onto which a formal network could not be trained. Hence, a large amount of effort have been put into direct SNN training.

### 5.2 Unsupervised Learning – Autonomous Extraction of Pattern in Datasets

The bio-inspiration and the supposed/anticipated consequent effectiveness of unsupervised learning led to consider the Spike Timing Dependent Plasticity (STDP) algorithm as a promising mean of training a SNN [30,120,170]. STDP is a Hebbian learning rule where the synapses' weight update depends only on the relative timings of pre- and post- synaptic neuron spikes, as depicted in Fig. 10. If the post-synaptic neuron fires shortly after the pre-synaptic one, a causality is expected between the two and hence their synaptic connection is strengthened. Inversely, if the first to fire is the post-synaptic one, then the relation is weakened. However, unsupervised learning via STDP does not allow multi layered networks to reach high accuracy [131]. That is mostly due to the close locality of Hebbian learning, where each layer adapts to the output of the preceding layer, without coordination between them to specify the transformation input to output of the full network. So, different strategies have been used to improve the accuracy of STDP trained networks: (1) add supervision to the training, such as reinforcement learning [45,131,132], (2) pre-process the input data [87,88], (3) modify the learning rule or the layer communication scheme [78,179]. These strategies allows to increase the performance on classification tasks, but they do not reach the state of the art with respect to both SNNs trained with supervision and usual ANNs trained with backpropagation.

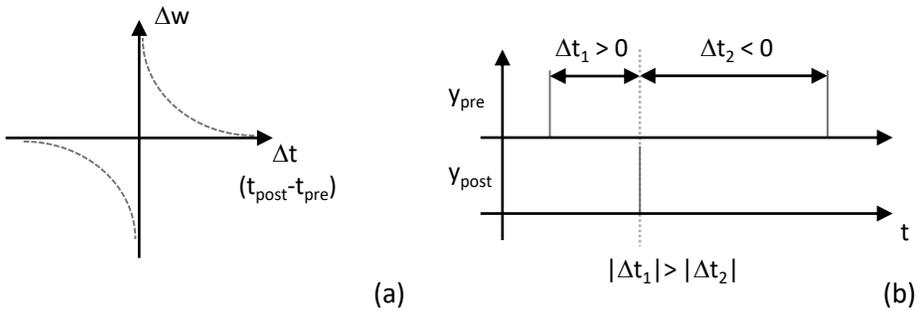

Fig. 10. Spike Timing Dependent Plasticity (STDP) rule illustration. (a) Diagram of the relation between weight update $\Delta w$ and relative timings of post- and pre-synaptic spikes $\Delta t$. (b) Chronograms illustrating the spike timings and associated weight updates.

Moreover, researchers are discussing the fact that a trained neural network is not able to extract semantic from the learnt data, but only statistically frequent features [82]. As a confirmation of this, Mozafari et al. [132] have shown that STDP allows to extract features occurring frequently, but not the ones specific to the targeted application. Therefore, even if, Kheradpisheh et al. [88] could train a convolutional SNN to an accuracy of 98.4% on MNIST, using pre-processing layers and a simplified version of STDP on the convolutional layers, it is unclear whether unsupervised learning in itself is suited for classification tasks.

We argue that such learning rules should be considered as a research tool or pre-processing method, to autonomously extract relevant patterns in complex inputs or to get a better understanding on what to focus on. For example, [102] realized an SNN training in two times by first using STDP and then with backpropagation based on converted spiking neurons outputs from spikes to activations values. Such methods could thus facilitate exploitation of complex data where spatio-temporal or spectro-temporal features are essential such as spike-sorting [16,57,188] or artificial finger haptic [152,203].

## 5.3 Supervised Learning – On the Road to Event Driven Machine Learning

While unsupervised learning is not efficient enough for high accuracy SNN training, the task of finding an efficient supervised learning rule remains. Several SNN-specific learning algorithms or technics [21,45,46,126,144,178,197] have been proposed, among which are the well-known ReSuMe [144], SPAN [126] and Chronotron [46]. These works follow different paths: (1) some are based on STDP and enable supervised training [45,144], (2) one analytically computes the weights of the network [178], (3) others approximate the behavior of spiking neurons to enable gradient computation from an error function [126,197].

The authors of [21] studied the possibility to backpropagate the gradient in SNN, and an increasing amount of recent publications refer to its implementation [42,69,81,104,129,190,193,199]. What limits backpropagation with spiking units is the non-derivable mathematical expression of neurons and synapses. Indeed, spikes are discontinuities and are thus non-derivable. To overcome this difficulty, several research groups explore various approaches. Lee et al. [104] simply consider that non-continuity is negligible, and show that ignoring it has only a small impact on final network accuracy. In [42], Esser et al. introduce side functions derivable to enable backpropagation without modifying the non-continuous units. This strategy is similarly used by Zenke et al. [199]. Others [129,193] exploit the continuity of models of neurons and synapses relying on time coding to enable derivation and computation of the gradient. Finally, some groups [69] design differentiable models of networks to enable gradient evaluation. These works show good accuracy on the MNIST and N-MNIST datasets, which are the more common datasets for SNN evaluation, but are still behind the conventional ANNs for similar network topologies. However, Kulkarni et al. [99] have recently shown training a SNN with better accuracy than its ANN counterpart, but only the final layer was trained with spiking model. Jin et al. [81] have also shown impressive statistics on these datasets by implementing backpropagation throughout a full SNN. They evaluate local gradients, at the "micro level", from the train of spikes of each neuron, and regard these spike traces as some sort of activations, which allows them to realize layer-to-layer gradient descent, at the "macro level", in a fashion similar to the classical formal approach. It allows them to achieve an accuracy of 99.49% on MNIST with a convolutional spiking network, which is, to the best of our knowledge, the best result demonstrated with a SNN.

Interestingly, Wu et al. [190] have introduced a framework for back propagating the gradient both along the depth and time dimensions of a spiking network, which allows the training to take into account the evolution of the data. This algorithm may enable a SNN to develop a better understanding of event-driven datasets by including the time dimension of the data as part of the searched patterns.

Some argue that using less bio-plausible models may allow deriving efficient learning rules while leveraging the sparse activity of spiking neurons [112]. Such philosophy is also present in the work of Yin et al. [193]. Indeed, they propose two different models of neurons, each adapted to tasks requiring computational elements with different degrees of bio-plausibility, and use them for different tasks. They propose two models of neurons allowing them to implement backpropagation of the gradient, a simple one that do not use time as a computational primitive, and another model which does.

To sum up, supervised learning of SNN, more specifically backpropagation of the gradient, is starting to enable spiking networks to reach accuracies equivalent to the ones of formal neural networks. Some models are hardware friendly and already enable low power and high accuracy inference [193]. Also, researchers [190] are working on solutions to include time evolution in the evaluation of the gradients for backpropagation, which allow to train the networks on event-driven datasets without any conversion required. Hence, it may be a matter of time before the SNNs finally deliver their full potential.

## 5.4 On-Chip or Off-Chip Training?

Whether to implement training on-chip or off-chip should be related to the considered application. If the objective is to design a general accelerator for machine learning, obviously the chip should allow on-chip training [24]. Now, if the purpose is to perform a unique machine learning task on embedded low power hardware, an off-chip learning, potentially power consuming, can be realized only once, after what the resulting network is programmed on the chip. At that point, one could argue that, in some cases, the system should need to adapt to its sensing environment while operating, what is referred to as on-line learning. One solution is to enable off-chip training in between operation times, and update/fine tune the system during inactive or loading time. However, this last statement still brings a lot of drawbacks: for example it requires to add a memory which would save the input data acquired during operation. Also on-line learning is still under huge research because machine learning currently has the major drawback of catastrophic forgetting, that is a trained network cannot learn a new task without losing accuracy on its previously learnt task [8,47,55].

The locality of STDP is an argument in favor of on-chip learning, because few or no information is transmitted between neurocores to enable weight update based on spike timing. The neurocore can store, with a counter, the time at which the pulse arrives and compares it to the firing time of the post-synaptic neuron. However, on-chip STDP, requires specific circuitry [92], and/or addition of memory to the neurocore, which may reduce power efficiency and increase the chip area. Nevertheless, some groups work on modified STDP rules to reduce its computational cost [78], and we have seen some demonstrations of on-chip STDP implementation with small hardware overhead [114,201], which may still contribute as an argument in favor of embedding spiking neural networks.

Finally, if we want to apply supervised learning, these algorithms require either complex neurons and synaptic models [126,144,197], or floating point values communication of gradient between layers, and thus between neurocores [104,193], which makes their hardware implementation impractical. Moreover, if weight update is performed on-line, i.e. during inference, feedforward operation must be paused for learning, which adds an operational delay to the system [201].

## 6 Hardware Implementations: Comparisons and Discussions

The question "how to compare neural network accelerators" is still unanswered. We believe that the comparison should be hierarchized in the following way. First, the type of neural network to be accelerated, i.e. ANN or SNN, should be differentiated, to avoid biasing one versus the other, as discussed in section 4.3. Then, one must consider the application. An embedded accelerator, with restricted power budget, cannot be compared to a general deep learning processor. They do not have the same constraints, in terms of area, power, flexibility, etc. Finally, one should decide which criteria are relevant for the targeted application.

For example, a general machine-learning accelerator focuses on speed of operation, flexibility of design to adapt to various topologies, and finally power to enable very large-scale algorithms evaluation. On the other hand, an embedded processor should focus on power consumption, accuracy of the mapped network and die area. Both devices perform the same operation, but their main constraints are different. In the next sections, we give an overview of both general SNN accelerators and application-specific small scale SNN processors.

### 6.1 General Neural Network Accelerators

Evaluation of large-scale neural networks of various topologies requires the accelerator to be highly configurable. The well-known large-scale architectures, TrueNorth [122], Neurogrid [15], BrainscaleS [160], Loihi [38], and SpiNNaker [49], adopt different characteristics to emulate networks of spiking neurons. We report the reader to Furber S. [48] for an advanced analysis of these designs.

SpiNNaker [49] uses a network of CPUs tightly connected to local memory, on a single chip. It is the most reconfigurable of the four, but it is not as energy efficient, neither as fast, as the others, especially if complex models of neuron and synapses are simulated. BrainScaleS [160] is implemented as several wafers interconnected together, each wafer consisting of several HiCANN neurocores. It targets the emulation of brain sized neural networks with precise biological neural behavior, at accelerated time. Neurogrid [15] is a SNN evaluator designed for sub-threshold analog electronics operation. It operates in real time, and emulates a few bio-realistic mechanisms. Finally, TrueNorth [122] is a neuromorphic chip implemented in digital electronics. It targets very low power large scale networks evaluation, and implements crossbar array with limited weight values representation and time multiplexed neuron updates.

These four chips are considered as a big step in the advance of spiking neuromorphic processors, mostly with the aim of mimicking biology, and TrueNorth targeting low power machine learning with spiking operators.

Recently, Intel has put a step in the balance with the Loihi processor [38]. It is a digital processor targeting flexibility of operation for large scale spiking neural network evaluation. In terms of functionality, they are at the frontier between bio-mimicry and machine learning with SNNs. It integrates on-chip learning with various learning rules implementable, complex neuron models, several information coding protocols, among others. It thus enables the emulation of many different algorithms. The authors target acceleration of the research on SNN performance. Table 1, inspired from [48], describes relevant characteristics for large scale system comparisons.

Table 1. Big Ones Characteristics

| Processor | BrainScaleS [160] | Neurogrid [15] | TrueNorth [122] | SpiNNaker [49] | Loihi [38] |
|---|---|---|---|---|---|
| Implementation | Analog | Analog | Digital | Digital | Digital |
| Time | Discretized | Real Time | Discretized | Discretized | Discretized |
| Neuron Update | | Real Time | Time MUX | Time MUX | Time MUX |
| Synapse Resolution | 4b | 13b shared | 1b | Variable | 1 to 64b |
| Bio-Mimicry | Not Configurable | Not Configurable | Limited to LIF | Configurable | Configurable |
| On-Chip Learning | STDP only | No | No | Yes | Yes |
| Network on Chip | Hierarchical | Tree Multicast | 2D Mesh Unicast | 2D Mesh Multicast | 2D Mesh Unicast |
| Neurons per Core | 8 to 512 | 65e3 | 256 | ~1e3 | max. 1024 |
| Synapses per Core | ~130k | 100e6 | 65k x 1b | ~1e6 | 16000 x 64b |
| Cores per Chip | 352 (wafer scale) | 1 | 4096 | 16 | 128 |
| Chip Area (mm$^2$) | 50 (single core) | 168 | 430 | 102 | 60 |
| Technology (nm) | 180 | 180 | 28 | 130 | 14 (FinFET) |
| Energy/SOP (pJ) | 174[a] | 941[a] | 27[b] | 27e3[c] | min. 105.3[d] |

[a]calculated as: *total power/(spike per second × number of synapses)*, data from [15]

[b]calculated as: *total power/(number of neurons × firing frequency × number of synapses)*, data from [122]

[c]calculated as: *total power/number of connections per second*, data from [49]

[d]calculated as: *active neural update + synaptc operation + within tile transmission*, data from [38]

## 6.2 Low Power Spiking Machine Learning

For what concerns application-specific accelerators, we focus here on the recently proposed image recognition processors. Visual classification is currently a major field of machine learning. It is so because of the potential applications in autonomous cars, robots, drones, and others. Therefore, many SNN accelerator designers focus on realizing image classifier accelerators. Spiking hardware proofs of concept are often benchmarked on the MNIST digit recognition dataset, even though it is a formal dataset. We believe it is so because it allows to train shallow SNNs, with various learning rules, and still obtain relatively high accuracies. Hence, we base our comparison on results obtained on this dataset. Nevertheless, we still argue that there are two major problems related to this dataset. First, it requires conversion from frame-driven to event-driven information, whose protocol may vary between benchmarked works. Second, it does not contain temporal information in itself, so it may not be adapted to exploit the full computational capacity of SNNs. Moreover, around 0.2% error on MNIST was already achieved in 2003 using a formal neural network [27], whereas the currently demonstrated best results using spiking networks are above 0.5% error [81,155]. Hence, we argue that researchers in the field should select a dataset onto which SNN accelerators could be compared fairly, where timing information is relevant, and no input conversion is required. Several event-driven datasets obtained with bio-inspired image sensors have already been proposed [18,125,169,204].

Recently, few groups have presented near ideal accuracies on visual datasets thanks to the development of powerful learning rules or conversion methods [81,88,104,155], but did not evaluate their algorithms on dedicated hardware. To perform operations on specialized processors, the algorithms are usually modified, to better fit the circuit capabilities, even if the hardware and software are co-designed. We thus focus on the works which resulted into neuromorphic circuit evaluation of SNN on MNIST [34,115,130,193,201]. We exclusively consider hardware-related data to give a comparison overview of the latest SNN accelerators targeting low power and potential embedding (Table 2). Moreover, for the sake of fairness, we add similar data measured on a very efficient formal ANN processor [189] in the last column of this table.

Yin et al. [193], in the SNN hardware category, reach the highest accuracy at smallest energy consumption for equivalent area. They train their network directly with spikes, off-chip, using a modified version of LIF neurons that enable backpropagation via Straight Through Estimator (STE) gradients. Through simulation of a spiking convolutional neural network using the same computational units, they reach 99.40% accuracy, which puts them slightly behind the state of the art [81,155]. However, Whatmough et al. [189] propose an ANN accelerator that realizes near identical accuracy (0.34% less) for 2× less energy per classification, but with more than 3× increase in area. Nevertheless, Yin et al. [193] report that, by reducing their accuracy by less than 0.2%, they can realize 3× gain energy at iso-accuracy with respect to Whatmough et al. [189]. It illustrates the ability of SNNs to achieve impressive energy accuracy tradeoffs. Interestingly, they can reach such efficiency because their hardware implementation is designed specifically for the SNN topology it evaluates. It thus lacks flexibility but trades it for superior energy efficiency.

Also, each of the presented processors use different strategies to improve its efficiency, such as weight quantization (WQ). However, it remains important to consider both hardware and software performances when dealing with energy efficiency improvement. For example, Zheng et al. [201] reduced the size and complexity of their topology in order to enable SNN evaluation at low circuit cost, but it induced an important accuracy loss. Indeed, before simplification, their network reached an accuracy of 97.8%, and after down-sampling the input data, evaluation with approximate computing, and other methods, they reduced their accuracy to around 90% on the same dataset.

We must specify that the data given for the chip of Chen et al. [34] are actually obtained through evaluation of a BNN onto the neuromorphic hardware. They present a way of evaluating formal binary networks on neuromorphic hardware with LIF neurons. Their threshold is set at zero, with an infinite leak rate, and the weights are quantized to either 0 or 1. Nevertheless, as proof of concept, they also test their hardware for SNN evaluation using on-chip STDP learning for image reconstitution, such as in [96].

Therefore, spiking neuromorphic hardware allows to perform MNIST digit recognition task for hundreds of nanojoules at more than 96% accuracies. They usually employ energy/accuracy tradeoff technics to improve hardware efficiency while losing a small percentage of accuracy. SNNs are thus low power, and can reach equivalent results on MNIST at lower power than the current state of the art ASIC for energy efficient ANN evaluation. Nevertheless, we argue that this dataset is not suited for SNN evaluation given that it does not allow these networks to exploit their full computational capability.

Table 2. Small Scale, Low Power Accelerators Comparison

| Processor | Yin et al. [193] | Mostafa et al. [130] | Zheng et al. [201] | Chen et al. [34] | Whatmough et al. [189] |
|---|---|---|---|---|---|
| Year | 2017 | 2017 | 2018 | 2018 | 2017 |
| Network Type | SNN | | | | ANN |
| Technology | Simulated 28nm | FPGA | Simulated 65nm | 10nm FinFET | 28nm |
| Implementation | Digital | Digital | Digital | Digital | Digital |
| Training Type | Backpropagation Off-Chip | Backpropagation Off-Chip | Modulated STDP On-Chip | Formal BNN Off-Chip | Formal Backpropagation |
| Coding Scheme | Rate | ItL[a] | Rate | Rate | NR |
| Accuracy on MNIST | 98.7 | 96.08 | ~90 | 97.9[b] | 98.36 |
| Spikes per Classification | NC | 135 | NC | ~130k | NR |
| Network Topology | FC 3 Layers | FC 3 Layers | FC 3Layers | FC 4 Layers | FC 5 Layers |
| Number of Neurons | 1306 | 1394 | 316[c] | 2330 | 1562 |
| Input Conversion Mode | Bernoulli Spike Probability | Binarized ItL | Continuous Values [200] | / | NR |
| Energy per Classification | 773nJ | NC | 1.12uJ | 1.7uJ | 360nJ |
| Chip Size | 1.65mm$^2$ | NC | 1.1mm$^2$ | 1.72mm$^2$ | 5.76mm$^2$ |
| Energy Efficiency | WQ[d] | WQ + Single Spike per Neuron + Approx. Multiplication (decay) | WQ + Approx. Division (training) + Random Refractoriness | WQ + Sparse Connect. + Stochastic Dropping | WQ + Approximate computing + Zero Guarding + Voltage Scaling |

[a]Intensity to Latency conversion

[b]Evaluation realized with a binary neural network on neuromorphic spiking hardware

[c]MNIST data down sampled from 28x28 to 16x16

[d]Weight Quantization

[e]NC=Not Communicated, NR=Not Relevant

# 7  WHAT COMES NEXT?

Throughout this survey, it became apparent that algorithm and hardware performances are interwoven. We would like to give an overview of the ways SNNs performances could be improved, from both software and hardware perspectives.

## 7.1  Algorithm Improvements

From a software point of view, SNNs still have an important improvement margin. Backpropagation of the gradient for supervised learning has recently delivered state of the art results [81,104]. It is, however, not compatible with every kind of networks. It requires modification of neuron models and thus still limits its scope of applications among SNN topologies. Nevertheless, it allows reaching good results and we may soon see machine learners start working with deep SNNs.

Moreover, information coding in SNN can allow reducing the total number of operation for a dedicated task on top of being intrinsically sparse. It is thus no surprise that an increasing number of works report implementing time coding SNNs [129,132,154,196]. Nevertheless, precise timing operations are still limited because of the lack of efficient training methods. In [131,132] they employ reward modulated STDP to train the network, but the learning rule does not enable multi-layer training. Rueckauer et al. [154] convert an ANN to a SNN using temporal encoding unit. They can thus train deep networks, but the conversion induces a loss in accuracy, with an error on MNIST of 1.04% and 1.43% before and after conversion respectively. Hesham Mostafa [129] uses backpropagation directly with spikes, but the reached accuracy on MNIST of 97.55% does not compete with a trained ANN. So, even if temporal coding is appealing thanks to the low number of spikes required to operate, algorithms based on it do not yet allow to reach accuracies comparable to state of the art algorithms, both formal and event-driven. Another way of reducing the number of spikes when using rate coding is to use adapting neuron models, which permits to maintain a low spike frequency while enlarging the dynamic range of the units. Zambrano et al. [198] demonstrate that using such units allow to reach higher accuracies than traditional rate coding SNNs while reducing the total number of operations per classification.

To sum up, SNN algorithms are very interesting because they allow both complex modelling of bio-plausible dynamics and machine learning tasks. The designers can adapt the complexity of their model to the targeted application. However, nowadays, the scope of available solutions is large and not one of these stand out of the crowd for machine learning tasks. Once again, this may be due to the fact that the benchmarks employed are not suited for fully exploiting SNN capabilities, given that researcher usually compare themselves to common ANNs.

## 7.2  Hardware Implementations

In terms of hardware, two fields are starting to stand out, the crossbar arrays with NVM devices and 3D integration. As we have already discussed in section 4.5, the crossbar arrays are quite promising for fast and low power neuromorphic architecture design. They are still at the state of research but promising results have already been demonstrated [5].

3D integration technology brings the advantages of high bandwidth, shorter interconnection designs, and potential high parallelism [140]. It enables interconnecting circuits on more than a single plan, with vertical wiring of the plans. Many works have already consider leveraging 3D technology to improve neuromorphic computing efficiency by implementing one layer by 3D plan [13,33], or by separating the memory and logic part on different tiers [33,89], or by maximizing parallel processing of analog/mixed signal designs with the use of crossbar arrays on the second plan [41]. Chang et al. [33] show that, in monolithic 3D, implementation of digital neuromorphic chip for formal processing is more performant if memory and logic are both distributed on several chips, which allows saving around 20% power with respect to the same 2D implementation.

However, in terms of speed, Kim et al. [89] show that stacking memory on top of logic plan can highly increase throughput with respect to the traditional memory-to-the-side implementation. Based on this, Amir et al. [6] are already envisioning sensors with integrated DNN computations at low footprint. Also, some groups [41,84] have proposed to exploit the Through Silicon Via technology process constraints on the density of interconnects to design analog neuron models with reduced capacitance footprint. Therefore, we see works dealing with formal ANN processors whose throughput and power consumption are increased and decreased respectively, or works studying the benefits of 3D for a parallel event-driven architecture. However, three-dimensional circuits is not a mature technology yet [12,158]. It brings several design constraints and comes at an important process cost. Moreover, because the AER protocol already allows low power communication between neurocores, further work is required to see how far 3D technology can improve the performances of SNN accelerators.

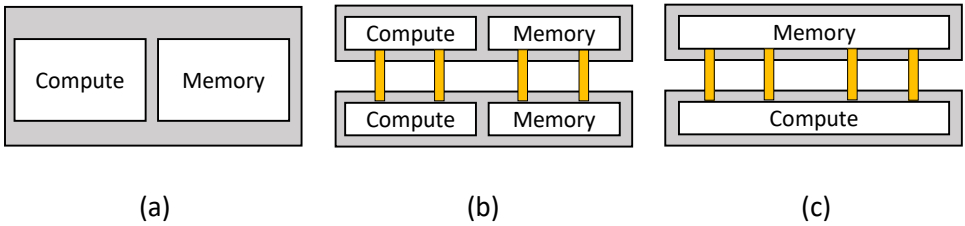

Fig. 11. Comparison of a traditional 2D implementation of a neurocore and novel 3D designs. (a) Schematic of a 2D design with a computing part and the memory aside of it. (b) 3D design with both algorithmic and memory circuitry in each tiers. (c) Another three dimensional implementation with a layer dedicated to memory and another to the computing elements.

Crossbar arrays and three-dimensional circuits are not the only trail of hardware improvement. For example, Morro et al. [128] have reported replacing a part of their ASIC, initially designed with traditional digital gates, by neuromorphic hardware. This sort of mixed neuromorphic integration is probably relevant for other applications. Yousefzadeh et al. [195] developed a chip for evaluating formal ANNs that use event-driven communication between layers, a topology they call "Hybrid Neural Network". It requires circuits to convert asynchronous event-driven information into frames, and inversely. They employs the AER protocol between layers, where information is encoded with 4 bits, thus guaranteeing minimal AER bus width overhead. For each non-zero activation, a single word packet is sent through the asynchronous NoC, which is sufficient to transmit the full information from neuron to neuron. They reach a relatively low accuracy of 97.09% on MNIST, for 7uJ/frame with an implementation on FPGA, which is below the state of the art (see section 6.2). However, further optimization and dedicated silicon implementation may allow performing better. It is an interesting approach in the sense that this technology could combine the advantages of both SNN and ANN while mitigating their drawbacks.

## 8  CONCLUSION

Neuromorphic computing is at an early stage, yet it is already a major field of research. In this survey, we focused on the hardware implementation of spiking neural networks and presented the state of the art. Neuromorphic processors for SNN acceleration can be designed for both deep network evaluation, where large scale circuits are implemented, and embedded applications, employing smaller scale accelerators. In section 4 we discussed the fact that hardware requirements are minimized thanks to intrinsic properties of SNNs. Moreover, algorithmic modifications, such as weight quantization or stochastic computing, help to reduce hardware resource requirements, in terms of memory and computation, which is beneficial for embedded operation. However, it may lead to a significant loss in accuracy of the network if those methods are not applied carefully, and formal networks can also leverage most of these technics. We observed that designers targeting machine learning applications are increasingly using digital implementations, which shows a cutoff point with the initial philosophy around hardware implementation of spiking neural networks which focused on bio-emulation in analog electronics. We suggest that it may be due to the fact that, up to now, bio-plausibility is not associated with better algorithmic performances, and many works are thus implementing simplified versions of neuron and synapse models, which are suited for digital evaluation.

This work also discussed the current trend in the community regarding benchmarking with the MNIST dataset, and we argued that it might not allow to fully exploit the SNN capabilities. Designers should prefer benchmarks dedicated to SNN evaluation, where temporal information is inherent to the dataset. Nevertheless, based on this dataset, current state of the art low power SNN accelerators are slightly better than ANN ones in terms of energy and area. It is important to note that the ANN reference of table 2 is a programmable accelerator able to manage several small topologies, whereas the accelerator of Yin et al. [193] has been specifically designed to implement a single network topology. We concede that this design strategy may not be suited for any applications, it is however very efficient.

Finally, we briefly presented the ongoing research on the evaluation of neural networks on crossbar array and 3D integration technology. Those may enable efficiency improvement of neuromorphic hardware, however crossbar arrays with memristive devices still do not meet the requirements and 3D technology is at a very early stage of research.